\documentclass[conference]{IEEEtran}
\IEEEoverridecommandlockouts
\def\BibTeX{{\rm B\kern-.05em{\sc i\kern-.025em b}\kern-.08em
    T\kern-.1667em\lower.7ex\hbox{E}\kern-.125emX}}

\usepackage{booktabs}
\usepackage[accsupp]{axessibility}
\usepackage[pagebackref,breaklinks,colorlinks,citecolor=red]{hyperref}
\usepackage{orcidlink}

\usepackage{amsmath,amsfonts,bm}

\def\eqref#1{equation~\ref{#1}}

\def\1{\bm{1}}

\DeclareMathAlphabet{\mathsfit}{\encodingdefault}{\sfdefault}{m}{sl}
\SetMathAlphabet{\mathsfit}{bold}{\encodingdefault}{\sfdefault}{bx}{n}

\DeclareMathOperator*{\argmin}{arg\,min}

\usepackage{comment}
\usepackage{wrapfig}
\usepackage{caption}
\usepackage{subcaption}
\usepackage{flushend}
\usepackage{balance}
\usepackage{lineno}
\usepackage{amsmath,amsfonts}

\usepackage{algorithm}
\usepackage{algorithmic}
\usepackage{hyperref}
\usepackage{graphicx}
\usepackage{textcomp}
\usepackage{xcolor}
\usepackage{url}
\usepackage{float}
\usepackage{multirow}
\usepackage{multicol}
\usepackage{color}
\usepackage{bm}
\usepackage{bbm}
\usepackage{enumitem}

\usepackage{pifont}

\newcommand{\bX}{\mathbf{X}}

\newcommand{\bA}{\mathbf{A}}

\newcommand{\bZ}{\mathbf{Z}}

\newcommand{\bH}{\mathbf{H}}
\newcommand{\bh}{\mathbf{h}}

\newcommand{\bz}{\mathbf{z}}

\usepackage{array}
\newcolumntype{L}[1]{>{\raggedright\let\newline\\\arraybackslash\hspace{0pt}}m{#1}}
\newcolumntype{C}[1]{>{\centering\let\newline  \\\arraybackslash\hspace{0pt}}m{#1}}
\newcolumntype{R}[1]{>{\raggedleft\let\newline \\\arraybackslash\hspace{0pt}}m{#1}}

\author{
\IEEEauthorblockN{
    Song Wang\IEEEauthorrefmark{2}, Xiaodong Yang\IEEEauthorrefmark{3}, Rashidul Islam\IEEEauthorrefmark{3}, Huiyuan Chen\IEEEauthorrefmark{3}, Minghua Xu\IEEEauthorrefmark{3}, Jundong Li\IEEEauthorrefmark{2}, Yiwei Cai\IEEEauthorrefmark{3}
}
\IEEEauthorblockA{\IEEEauthorrefmark{2}\textit{University of Virginia}, Charlottesville, USA, \{sw3wv, jundong\}@virginia.edu}
\IEEEauthorblockA{\IEEEauthorrefmark{3}\textit{Visa Research}, USA, \{xiaodyan, raislam, hchen, mixu, yicai\}@visa.com}
}

\begin{document}

\title{Enhancing Distribution and Label Consistency\\ for Graph Out-of-Distribution Generalization

}

\maketitle

\begin{abstract}
To deal with distribution shifts in graph data, various graph out-of-distribution (OOD) generalization techniques have been recently proposed.
These methods often employ a two-step strategy that first creates augmented environments and subsequently identifies invariant subgraphs to improve generalizability. Nevertheless, this approach could be suboptimal from the perspective of \textit{consistency}. First, the process of augmenting environments by altering the graphs while preserving labels may lead to graphs that are not realistic or meaningfully related to the origin distribution, thus lacking \textit{distribution consistency}.
 Second, the extracted subgraphs are obtained from directly modifying graphs, and may not necessarily maintain a consistent predictive relationship with their labels, thereby impacting \textit{label consistency}.
%
In response to these challenges, we introduce an innovative approach that aims to enhance these two types of consistency for graph OOD generalization. We propose a modifier to obtain both augmented and invariant graphs in a unified manner. With the augmented graphs, we enrich the training data without compromising the integrity of label-graph relationships. The label consistency enhancement in our framework further preserves the supervision information in the invariant graph. 
We conduct extensive experiments on real-world datasets to demonstrate the superiority of our framework over other state-of-the-art baselines.
\end{abstract}


\section{Introduction}
Graph data is prevalent in various applications, such as molecular property predictions and financial analysis~\cite{tang2008arnetminer,mcauley2015inferring,zhou2019meta,bojchevski2018deep,tan2022graph, wang2023contrast}. Recently, techniques in graph machine learning have significantly advanced. For example, Graph Neural Networks (GNNs) have been developed to effectively handle graph data~\cite{kipf2017semi,velivckovic2017graph,zhou2020graph}. GNNs learn representations of individual nodes by aggregating information from their neighboring nodes~\cite{chang2015heterogeneous,hamilton2017inductive,xu2018powerful,wang2022glitter}. Despite the notable success of graph machine learning, a significant challenge remains. GNNs typically assume that the training data follows the same distribution as the test data. However, this assumption often does not hold true in real-world scenarios~\cite{cao2016deep,miao2022interpretable,yu2023mind, wang2024safety}. Consequently, when the test data distribution differs significantly, the performance of GNNs trained on the training data deteriorates. This issue is commonly referred to as distribution shift~\cite{wu2022handling,you2018graphrnn,bianchi2021graph}.

Although numerous solutions to distribution shifts have been proposed for Euclidean data (e.g., images)~\cite{mansour2009domain,blanchard2011generalizing,muandet2013domain,beery2018recognition,recht2019imagenet,su2019one}, these techniques are challenging to apply to graphs due to their complex node connections~\cite{fakhraei2015collective,gui2022good, wang2022xfnc}. To address distribution shifts in graphs, recent methods have proposed generating augmented environments through data augmentation~\cite{krueger2021out,chang2020invariant}. By training on this augmented data, GNNs can learn the decisive information for classification while ignoring the spurious information of each environment, thereby achieving the robustness necessary to perform well on test data~\cite{albadawy2018deep,dai2018dark}.

However, such a strategy can be suboptimal, facing two crucial challenges: 
\ding{182} \textbf{\textit{Distribution Consistency}}: When generating augmented environments, the graphs are obtained via modification, which might diverge significantly from real-world examples. For example, in molecular property prediction, minor alterations to molecular structures can result in significant changes in chemical properties, thus disrupting the distribution consistency of augmented graphs~\cite{wu2022discovering}. This inconsistency arises because the augmented graphs may not accurately reflect realistic or probable molecular configurations.
\ding{183} \textbf{\textit{Label Consistency}}: The process of extracting invariant subgraphs, without careful consideration of how these changes impact label relevance, risks impairing the learned relationship between graph structures and labels. This is because the identified invariant subgraphs may not accurately reflect the labels they should belong to, which diminishes the preserved supervision information. 

To deal with these challenges, we propose a novel framework named \textbf{DLG}, which aims to enhance \textbf{\underline{D}}istribution- and \textbf{\underline{L}}abel-consistency for \textbf{\underline{G}}raph OOD generalization. Specifically, we design a novel graph modifier to produce both augmented and invariant graphs by sampling edges from the learned edge masks. Using such a strategy enables us to unify the processes of augmenting existing graphs and extracting invariant graphs by formulating them both as modifications, in order to enhance consistency from perspectives of distributions and labels. 
\ding{182} \textbf{\textit{Distribution Consistency Enhancement.}} To obtain augmented graphs during training while preserving distribution consistency, we propose to maximize the information between augmented graphs and existing graphs. In this manner, we could ensure the augmented graphs are informative and align with existing graphs to preserve distribution consistency. 
\ding{183} \textbf{\textit{Label Consistency Enhancement.}} To ensure the generated invariant graphs remain consistent across different domains regarding their labels, we explicitly enhance label consistency by ensuring that the extracted invariant subgraphs share maximal supervision information with the original graph $G$.  
In summary, our contributions are as follows:
\begin{itemize}[leftmargin=0.35cm]
        \item  [$\star$] \textbf{Investigation.} We investigate the challenges 
    of invariant learning in the problem of OOD generalization on graphs, with a particular emphasis on \emph{Distribution Consistency} and \emph{Label Consistency}. 
    \item [$\star$] \textbf{Design.} We develop a novel OOD generalization framework with two essential strategies: (1) a distribution consistency enhancement module that aims to achieve augmented graphs that exhibit better cross-domain consistency; (2) a label consistency enhancement module that ensures that the extracted invariant subgraph contains maximal decisive information for classification; 
  \item [$\star$] \textbf{Effectiveness.} We conduct extensive experiments on a variety of both graph-level and node-level graph OOD generalization datasets that cover both synthetic and real-world graph data. The results further demonstrate the superiority of our proposed framework.
\end{itemize}

\section{Related Works}
The challenge of out-of-distribution (OOD) generalization is to enable models to generalize to previously unseen test distributions after training on similar but distinct data~\cite{arjovsky2019invariant, ganin2015unsupervised}. Recent research has focused on invariant learning~\cite{li2018domain, ding2020graph}, which aims to maintain a stable relationship between inputs and outputs across different distributions. Current approaches typically achieve this by learning invariant representations~\cite{sun2016return} or invariant causal predictions~\cite{buhlmann2020invariance, heinze2018invariant}. Additionally, some methods seek OOD generalization by optimizing the worst-case group performance without predefined partitions in the training data~\cite{hu2018does, qian2019robust,sagawa2020distributionally}.


The issue of OOD generalization in graph data has gained increased attention recently~\cite{wu2023adversarial,sui2022adversarial}, driven by the prevalence of distribution shifts in various real-world scenarios~\cite{li2022learning}. Among these approaches, EERM~\cite{wu2022handling} was proposed for node classification and uses invariance principles to generate domains by maximizing loss variance across domains in an adversarial manner. For graph classification tasks, DIR~\cite{wu2022discovering} employs graph representations as causal rationales and generates additional distributions via interventional augmentations.
CIGA~\cite{chen2022learning} and DisC~\cite{fan2022debiasing} both propose to capture the causal information in subgraphs.
To address the limitations of CIGA and DisC in scenarios where the prior knowledge about the variance of causal and spurious information is unavailable, GALA~\cite{chen2023does} uses an assistant model to detect distribution shifts based on this variance. 

\section{Problem Formulation}
We first denote the set of training graphs as $\mathcal{D}_{tr}=\{(G_i,Y_i)|i\in[1,|\mathcal{D}_{tr}|]\}$. Here $G\in\mathcal{G}$ and $Y\in\mathcal{Y}$ are graph samples and their labels. 
We further denote each graph as $G=(\bA,\bX)$, where $\bA\in\mathbb{R}^{N\times N}$ and  $\mathbf{X}\in\mathbb{R}^{N\times d_x}$ represents the adjacency matrix and feature matrix, respectively. Here $N$ is the number of nodes in $\mathcal{G}$, and the $j$-th row vector of $\bX$ is the $d_x$-dimensional attributes of the $j$-th node.
Following existing works~\cite{chen2022learning,li2022learning,wu2022discovering}, the goal of OOD graph generalization is to learn a classification function
$f_c(\cdot)\colon\mathcal{G}\rightarrow\mathcal{Y}$ that takes a graph as input and predicts its label $Y$. The ultimate goal of OOD graph generalization is to learn the optimal classification function $f^*$ that could precisely predict the labels of graphs in the test set, i.e., $\mathcal{D}_{te}=\{(G^*_i,Y^*_i)|i\in[1,|\mathcal{D}_{te}|]\}$

\section{Methodology}

\begin{figure*}[!t]
	    \centering
	    \includegraphics[width=0.85\textwidth]{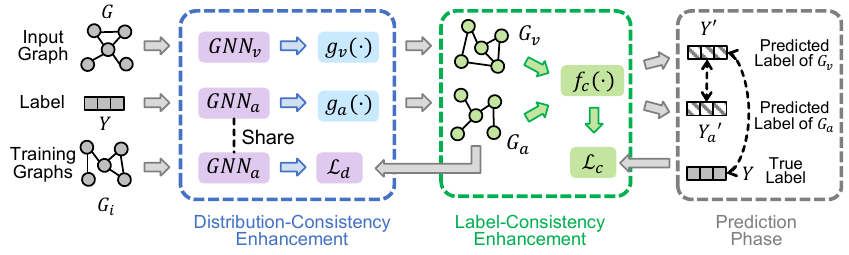}
	    \vspace{-0.05in}
\caption{The overall framework of DLG. Given a graph $G=(\bA, \bX)$, we feed it into $\text{GNN}_v$ to learn a graph representation, which will be used to generate an invariant graph $G_v$ for classification. Meanwhile, the input graph is also processed by $\text{GNN}_a$ to generate an augmented graph $G_a$, which shares the same label as $G_v$ but with different structures and edges. To enhance label consistency, we propose the classification loss $\mathcal{L}_c$ for optimization. To enhance distribution consistency, we design a loss $\mathcal{L}_d$ that leverages randomly sampled graphs from existing data to ensure $G_a$ aligns with distributions of existing data.}
\label{fig:illustration}
	\end{figure*}

We elaborate on our framework DLG for graph OOD generalization problem via enhancing distribution and label consistency. As illustrated in Fig.~\ref{fig:illustration}, our framework leverages a classifier $f_c(\cdot)$ and two graph modifiers $g_a(\cdot)$ and $g_v(\cdot)$, which generate augmented and invariant graphs, respectively. The encoder and modifiers are optimized based on our designed objectives for the two types of consistency. Note that our modifier only considers the edges in each graph, i.e., only modifying the adjacency matrix. That being said, we could potentially create edges that do not exist in the original graph.


\subsection{Modification}

Considering an input graph $G=(\bA, \bX)$ from the training dataset $\mathcal{D}_{tr}$, we first aim to learn its invariant graph $G_v=(\bA_v, \bX_v)$ that preserves label consistency for classification. Additionally, during training, we also generate an augmented graph $G_a=(\bA_a, \bX_a)$ that preserves distribution consistency. The augmented graph $G_a$ is also classified by the classifier $f(\cdot)$ with the same label as $\mathcal{G}$.  

In the following, we describe the process of obtaining the invariant graph $G_v$ and the augmented graph $G_a$ in detail. Particularly, we first learn the representations for modification:
\begin{equation}
  \bH_v= \text{GNN}_v(\bA, \bX)\quad \text{and}\quad\bH_a=\text{GNN}_a(\bA, \bX\oplus\mathbf{y}).
\end{equation}
Here $\text{GNN}_v$ and $\text{GNN}_a$ are two GNNs used in $g_v(\cdot)$ and $g_a(\cdot)$, respectively. $\bH_v$ (or $\bH_a$) represents the learned embedding matrix of $G_v$ (or $G_a$).  
Notably, for the input to $\text{GNN}_a$, we additionally incorporate the information from label $Y$ of $G$. This is because the augmented graphs are only used during training, which means its label information is always available. Moreover, with label information added, it also benefits our label consistency enhancement when used for training, as introduced in the following section. 
During the inference stage, the label information is unavailable, and thus we do not incorporate label $Y$ in $g_v(\cdot)$.

With the learned representations, we illustrate the detailed process of modification. As the process applies to both invariant and augmented graphs, here we use $\bH$ for simplicity. Particularly, given $\bH$, our target is to learn from edge masks $\{e_{i,j}|i,j=1,2,\dotsc, N\}$. Here each mask is in $[0,1]$, denoting the sampling probability of a specific edge. 
In general, a larger mask leads to a higher sampling probability, indicating that the edge is more likely to appear in the invariant graphs or augmented graphs. 

To learn the edge masks, we apply a Multi-Layer Perceptron (MLP) and calculate the dot product:
\begin{equation}
    e_{i,j}=\text{Sigmoid}\left(\left(\text{MLP}_e(\bh_{i})\right)^\top \cdot \text{MLP}_e(\bh_{j})\right),
\end{equation}
where $\bh_v$ denotes the representation of the $i$-th node, i.e., the $i$-th row vector in $\bH$. 
With the learned masks, we could perform sampling for edges. Nevertheless, such a sampling process is non-differentiable. To enable the optimization of our framework via gradient descent, we utilize the sigmoid variant of gumbel-softmax~\cite{jang2016categorical} to sample edges. By factorizing the modification process into sampling individual edges, we could avoid the extensive computational cost caused by the large graph space~\cite{sui2022causal}.

\subsection{Distribution Consistency Enhancement}\label{sec:dis}
Although we generate augmented graphs to expand the training data distribution, the strategy cannot guarantee that the generated graphs preserve consistency with existing graphs. In other words, the generated graphs may largely deviate from the distribution of existing data. Moreover, the labels of augmented graphs may also be different from the graph before augmentation. 

To deal with this, we propose to maximize the mutual information between augmented graphs $G_a$ and other existing graphs with the same label. Specifically, we first randomly sample a batch of graphs from all training data as the support set $\mathcal{S}=\{G_i|i=1,2,\dotsc,C\}$, containing one graph for each class. Here $C$ is the number of classes. Moreover, the label of $G_i$ is $y_i$, i.e., the $i$-th label in the label space $\mathcal{Y}$. In this manner, the objective of maximizing the mutual information between $G_a$ and $\mathcal{S}$ can be represented as follows:
\begin{equation}
    \max I(G_a; \mathcal{S})=\max \sum\limits_{i=1}^C p(G_a, G_i)\log \frac{p(G_a,G_i)}{p(G_a)p(G_i)}.
\end{equation}
Since the mutual information
$I(G_a; \mathcal{S})$ is intractable and thus infeasible to maximize~\cite{oord2018representation}, we transform the objective into a more accessible form:
\begin{equation}
     I(G_a; \mathcal{S})=\sum\limits_{i=1}^C p(G_a|G_i)p(G_i)\log \frac{p(G_i|G_a)}{p(G_i)}.
\end{equation}
This form is easier to optimize as it contains the term $p(G_i)$. As we randomly sample $G_i$ from each class, we  assume that the prior probability, i.e., $p(G_i)$, follows a uniform distribution and set it $p(G_i)=1/C$. According to Bayes’ theorem, the objective then becomes:
\begin{equation}
    I(G_a; \mathcal{S})=\frac{1}{C}\sum\limits_{i=1}^C p(G_a|G_i)\left(\log {p(G_i|G_a)}-\log\frac{1}{C}\right).
\end{equation}
To achieve the value of $p(G_a|G_i)$, we estimate it as $p(G_a|G_i)=\mathbbm{1}(Y=y_i)$, where $Y$ and $y_i$ are the labels of $G$ and $G_i$, respectively. Here $\mathbbm{1}(Y=y_i)=1$ if $G$ and $G_i$ share the same label; otherwise $\mathbbm{1}(Y=y_i)=0$. In this manner, considering that $\log(1/C)$ is a constant, the above objective can be further simplified as follows:
\begin{equation}
    \max I(G_a;\mathcal{S})= \max \frac{1}{C}\sum\limits_{i=1}^C \mathbbm{1}(Y=y_i)\log {p(G_i|G_a)}.
\end{equation}
To further estimate $\log p(G_i|G_a)$, we can define the probability of $G_i$ sharing the same label as $G$ (i.e., $y_i=Y$) according to the squared $\ell_2$ norm of the embedding distance. Specifically, we first apply the same $\text{GNN}_a$ on $G_i$ to obtain its representation:
\begin{equation}
            \bh_i=\frac{1}{|\mathcal{V}_i|}\sum\limits_{j=1}^{|\mathcal{V}_i|} \bH_i(j),\ \text{where}\ \bH_i= \text{GNN}_a(\bA_i, \bX_i\oplus \mathbf{y}_i),
\end{equation}
where $\bA_i$ and $\bX_i$ are adjacency matrix and feature matrix of $G_i$, respectively. $\mathbf{y}_i$ is the corresponding one-hot vector of the label of $G_i$. Based on the representation, the term $p(G_i|G_a)$ can be formulated as below, after normalization with a softmax function:
\begin{equation}
            \left.p(G_i|G_a)=\frac{\exp\left(-(\mathbf{h}_a- \mathbf{h}_i)^2\right)}{\sum_{j=1}^C\exp\left(-(\mathbf{h}_d-\mathbf{h}_j)^2\right)}
        \right.,
        \label{eq:prob}
\end{equation}
where $\bh_a$ and $\bh_i$ are graph-level representations of $G$ and $G_i$, respectively. Note that as $G_a$ is generated from $G$, here we use the graph representation of $G$ as $\bh_a$ to obtain the objective. If we further apply the $\ell_2$ normalization to both $\mathbf{h}_a$ and $\mathbf{h}_i$, we obtain $(\mathbf{h}_a- \mathbf{h}_i)^2=2-2\mathbf{h}_a\cdot \mathbf{h}_i$. 
Additionally, to enhance diversity, we still need to ensure the obtained $G_a$ is not excessively similar to $G_v$, even though they are of the same label. Thus, we modify the denominator in Eq.~(\ref{eq:prob}) to involve the graph-level representation of $G_v$, i.e., $\bh_v$. The objective becomes:
\begin{equation}
\begin{aligned}
     \mathcal{L}_d=&-\sum\nolimits_{i=1}^C \mathbbm{1}(Y=y_i)2\mathbf{h}_a\cdot \mathbf{h}_i- \\
&\log(\exp\left(2\mathbf{h}_a\cdot\mathbf{h}_v\right)+\sum\nolimits_{j=1}^C\exp\left(2\mathbf{h}_a\cdot\mathbf{h}_j\right)).
     \end{aligned}
           \label{eq:L_d}
\end{equation}
With this objective, we aim to ensure that the representation $\bH_a$ used for generating $G_a$ should be similar to existing graphs with the same label while distinct from other graphs with different labels, and more importantly, the original graph $G$. Consequently, we are able to broaden the distribution of the class to which $G$ belongs, through the use of the generated $G_a$, while also preserving the informativeness of $G_a$.




\subsection{Label Consistency Enhancement}\label{seq:label}
As our goal for the extracted invariant graph is to ensure label consistency, we strive to preserve as much useful information from the original graph $G$ as possible. Therefore, we formulate our objective as maximizing the conditional likelihood of $P(G_v|G, Y)$, representing the probability of the generated consistent graph $G_v$ conditioned given the original graph $G$. That being said, we aim to ensure that the generated $G_v$ share maximal supervision information with $G$ while ignoring unrelated information. Therefore, we obtain the classification losses:
$
    \mathbf{q}_v=f(G_v),\ \mathbf{q}_a=f(G_a),
$
and
\begin{equation}
    \mathcal{L}_c= -\sum_{i=1}^C\Bigl(\mathbbm{1}(Y=y_i) \log(q_v(i))+ q_v(i)\log(q_a(i))\Bigr),
    \label{eq:L_c}
\end{equation}
where $\mathbf{q}_v\in\mathbb{R}^C$ (or $\mathbf{q}_a\in\mathbb{R}^C$) is the probability that $G_v$ (or $G_a$) belongs to each class in the label space $\mathbb{Y}$, and $C$ is the number of classes. Moreover, $q_v(i)$ (or $q_a(i)$) is the $i$-th element in $\mathbf{q}_v$ (or $\mathbf{q}_a$). Here $\mathbbm{1}(Y=y_i)=1$ if the label $Y$ of $G$ is $y_i$; otherwise $\mathbbm{1}(Y=y_i)=0$. In this way, we enforce the class probability distributions of $G_v$ and $G_a$ to be close to each other while utilizing label information of $G$.

\subsection{Optimization}
In this subsection, we introduce the detailed optimization steps of DLG. As described in Sec.~\ref{seq:label} and Sec.~\ref{sec:dis}, we propose two losses to optimize our framework, i.e., $\mathcal{L}_d$ and $\mathcal{L}_c$. However, directly applying all two losses can be detrimental to the overall performance, as the modules in our framework maintain different objectives. Let us denote the parameters of $\text{GNN}_v$ and $\text{GNN}_a$ as $\theta_v$ and $\theta_a$, respectively. We apply both losses for optimization:
\begin{equation}
    \{\theta_v^*, \theta^*_a\}= \ \  \argmin_{\{\theta_v, \theta_a\}}  \ \  
 \alpha\mathcal{L}_d+(1-\alpha)\mathcal{L}_c,
            \label{eq:loss_c}
\end{equation}
where $\theta_v^*$ and $\theta_a^*$ are the optimal parameters for $\text{GNN}_v$ and $\text{GNN}_a$, respectively. $\alpha\in[0,1]$ is a hyper-parameter that controls the importance of $\mathcal{L}_d$.


Moreover, as the classifier is only involved in the classification loss, we optimize it with $\mathcal{L}_c$:
\begin{equation}
\theta_f^*=\argmin_{\theta_f} \mathcal{L}_c, 
        \label{eq:loss_d}
\end{equation}
where $\theta_f^*$ denotes the optimal parameters for the classifier $f_c(\cdot)$, respectively. In this manner, we can jointly optimize all modules in our framework in an end-to-end manner. 

\section{Experiments}
In this section, we evaluate our framework DLG on a variety of synthetic and real-world graph datasets for out-of-distribution generalization.

\subsection{Datasets and Settings}


\noindent\textbf{Graph-Level OOD Datasets.}
For experiments, we utilize a combination of synthetic and real datasets for OOD generalization on graph classification tasks. 
Specifically, we first consider graph-level datasets used in GALA~\cite{chen2023does}. The included datasets are as follows: \ding{182} TPG (Two-Piece Graph) datasets constructed based on the BA-2motifs~\cite{luo2020parameterized}, resulting in four variants of 3-class two-piece graph datasets. Each variant is controlled by a hyperparameter that adjusts the correlations between the label and the spurious information in the graph, ranging from $+0.2$ to $-0.2$.  \ding{183} Six Datasets from the DrugOOD benchmark~\cite{ji2022drugood}, which is centered on the challenging real-world task of AI-aided drug affinity prediction. The DrugOOD datasets include splits based on Assay, Scaffold, and Size from two categories: EC50 (denoted as EC50-) and Ki (denoted as Ki-). These datasets provide a diverse range of scenarios to evaluate the effectiveness of various methods in predicting drug affinities across different biological and chemical contexts. We further consider four graph-level datasets used in DIR~\cite{wu2022discovering}: \ding{184} \textbf{SP-Motif}~\cite{ying2019gnnexplainer} is a synthetic dataset, in which the degree of bias can be manually controlled. Each graph consists of a base (with structures from Tree, Ladder, and Wheel) and a motif (with structures from Cycle, House, and Crane). The label is entirely decided by the structure of the motif in each graph. We consider the average results of four bias degrees: $1/3$, $0.5$, $0.7$, and $0.9$. \ding{185} \textbf{MNIST-75sp}~\cite{knyazev2019understanding} is converted from the MNIST image dataset, and random noises are added to create distribution shifts. There exist 10 labels in the dataset. Moreover, the node features contain random noises in the test set. \ding{186} \textbf{Molhiv} (OGBG-Molhiv)~\cite{wu2018moleculenet,hu2020open,hu2021ogb} is a molecular property prediction dataset, and the distribution shift is based on different structures of molecules. \ding{187} \textbf{Graph-SST2 }contains graphs constructed from sentimental sentences, and the distribution shift originates from different partitions of graphs.
Note that due to the different properties of graph structures in these datasets, we follow DIR~\cite{wu2022discovering} to use different GNN backbones for various datasets. Among these four datasets, the SP-Motif dataset is synthetic, while the other three datasets are obtained from real-world data.
Our code is provided at \href{https://github.com/SongW-SW/DLG}{https://github.com/SongW-SW/DLG}.

\begin{table*}[t]
			\setlength\tabcolsep{8pt}
		\small
		\centering
  		\renewcommand{\arraystretch}{1.3}
\caption{The performance of various methods on TPG datasets with different degrees of bias. We report the results for test accuracy (in \%), and the best results are in \textbf{bold}.}
  \setlength{\aboverulesep}{0pt}
\setlength{\belowrulesep}{0pt}
\begin{tabular}{c|ccccccc}
\toprule[1.2pt]
{Dataset} & {TPG} & EC50 & Ki & SP-Motif & {MNIST-75sp} & {Graph-SST2} & {Molhiv} \\\hline
ERM & 61.77  $\pm$ 1.78 & 67.53 $\pm$ 1.45 & 73.54 $\pm$ 1.76 & 38.81 $\pm$ 1.61 & 12.71{ $\pm$ 1.43} & 81.44{ $\pm$ 0.59} & 76.20{ $\pm$ 1.14} \\
IRM & 61.38  $\pm$ 1.63 & 67.93 $\pm$ 1.28 & 73.56 $\pm$ 1.98 & 39.71 $\pm$ 2.11 & 18.62{ $\pm$ 1.22} & 81.01{ $\pm$ 1.13} & 74.46{ $\pm$ 2.74} \\
V-REx & 60.94  $\pm$ 1.98 & 67.70 $\pm$ 1.20 & 74.08 $\pm$ 1.86 & 39.04 $\pm$ 1.97 & 18.92{ $\pm$ 1.41} & 81.76{ $\pm$ 0.08} & 75.62{ $\pm$ 0.79} \\\hline
DIR & -- & --& --& 43.94 $\pm$ 1.70& 20.36{ $\pm$ 1.78} & 83.29{ $\pm$ 0.53} & 77.05{ $\pm$ 0.57}  \\
GIL & 64.50  $\pm$ 2.13 & 64.20 $\pm$ 3.18 & 73.69 $\pm$ 2.56 & 52.29 $\pm$ 2.96 & 21.94{ $\pm$ 0.38} & 83.44{ $\pm$ 0.37} & 79.08{ $\pm$ 0.54} \\
CIGA & 67.18  $\pm$ 1.85 & 68.18 $\pm$ 1.57 & 73.43 $\pm$ 2.71 & 69.86 $\pm$ 3.41 & 24.81{ $\pm$ 2.08} & 81.43{ $\pm$ 1.26} & 79.70{ $\pm$ 1.55} \\
GALA& 79.21 $\pm$ 1.34& 69.36 $\pm$ 1.51 & 76.16 $\pm$ 1.80 & -- &-- &--&-- \\\hline
DLG  & \textbf{79.69  $\pm$ 1.65} & \textbf{69.85 $\pm$ 1.84} & \textbf{76.55 $\pm$ 2.04} & \textbf{73.34 $\pm$ 2.84} & \textbf{31.22} $\pm$ \textbf{0.97} & \textbf{84.01} $\pm$ \textbf{0.89} & \textbf{{79.80} $\pm$ {0.60}} \\
\bottomrule[1.2pt]
\end{tabular}

\label{tab:results_two_piece}
\end{table*}

\noindent\textbf{Baselines.}
To evaluate our proposed framework DLG while comparing other state-of-the-art methods, we consider the following baselines: \ding{182} OOD methods proposed for Euclidean data (e.g. images), which include IRM~\cite{arjovsky2019invariant} and V-REx~\cite{krueger2021out}. \ding{183} OOD methods proposed for graph data, which include DIR~\cite{wu2022discovering},  GIL~\cite{li2022learning}, CIGA~\cite{chen2022learning}, and GALA~\cite{chen2023does}. 



\subsection{Results of Graph-Level OOD Generalization}

We conduct experiments for out-of-distribution generalization on two sets of datasets used in GALA~\cite{chen2023does} and DIR~\cite{wu2022handling}, respectively. We present the results of TPG (Two-Piece Graph), DrugOOD, and the four datasets used in DIR in Table~\ref{tab:results_two_piece}.  From the results, we have the following observations:
\ding{182} \textbf{DLG achieves the best result across most synthetic and real-world datasets.}
We observe that the performance of IRM and V-REx is less competitive, demonstrating that directly applying OOD generalization methods to graph data is insufficient. In concrete, these results indicate that DLG maintains superior generalizability compared to other baselines across various datasets.
\ding{183} \textbf{DLG exhibits superior performance on datasets with node features of fewer dimensions.} Among the datasets evaluated, the Graph-SST2 dataset uniquely preserves node features with a substantial dimension size of 768. In contrast, all other datasets maintain node features with sizes less than 10. This characteristic emphasizes the need to learn consistent representations while concurrently preserving their informativeness. Owing to the proposed informative loss integrated into our framework, DLG effectively retains informativeness within the learned representations, preventing the omission of valuable information when applied to datasets with limited node features.




		\begin{figure}[!t]
		\centering
\includegraphics[width=0.99\linewidth]{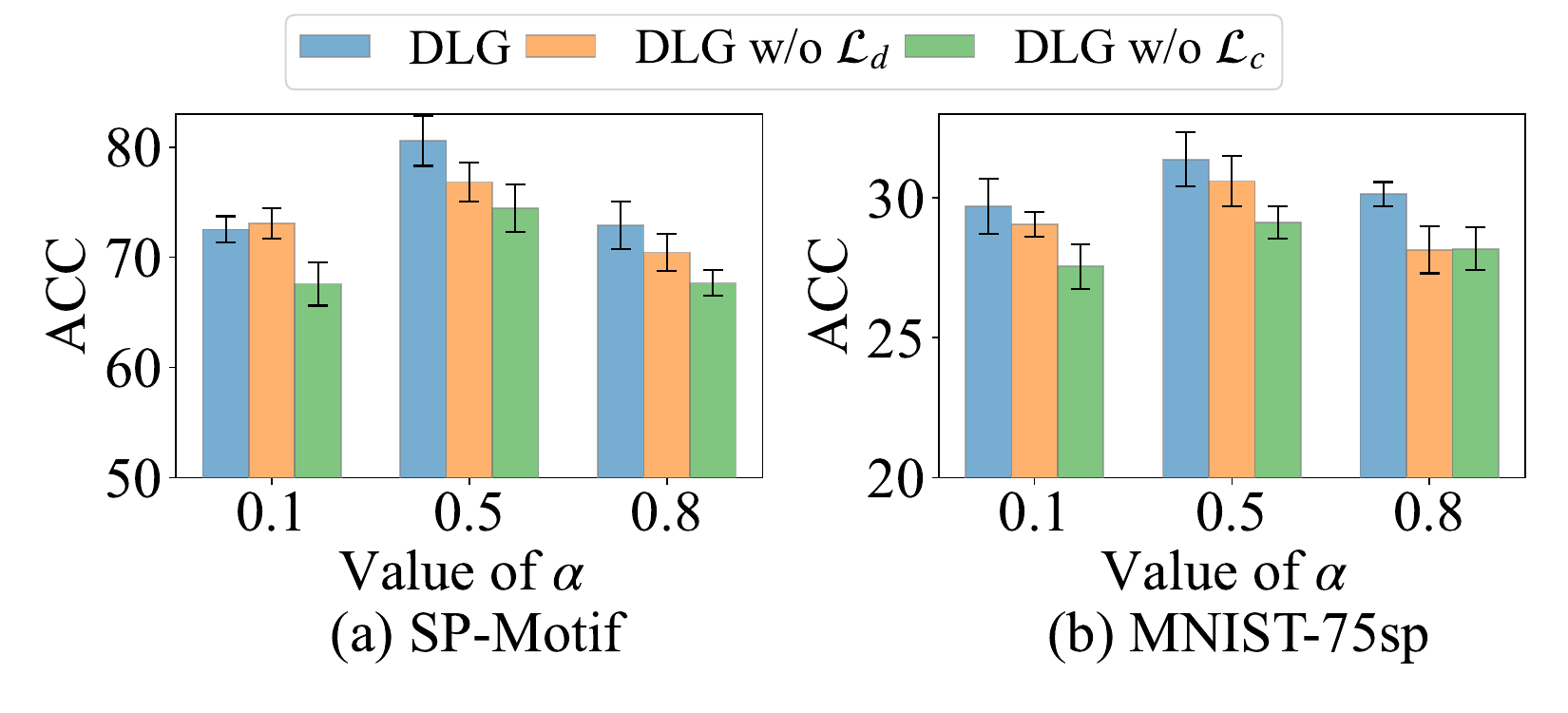}

		\caption{The performance of our framework DLG with different modules removed.  }
  \label{fig:ablation}

	\end{figure}

\subsection{Ablation Study of DLG}
In this subsection, we further present several ablation studies on various datasets to validate the significance of the two key designs within DLG. 
First, the diversity enhancement component is disabled by removing the loss $\mathcal{L}_d$, causing the framework unable to preserve distribution consistency. This variant is denoted as DLG w/o $\mathcal{L}_d$. Second, we discard the consistency enhancement by removing $\mathcal{L}_c$. This means that the augmented graphs and invariant graphs are not restricted by label consistency enhancement. We denote this version as DLG w/o $\mathcal{L}_c$. 
The results from our ablation study are illustrated in Fig.~\ref{fig:ablation}. From these results, we observe that DLG outperforms all other variants, which validates the effectiveness of the two modules in our framework. 
First, integrating the distribution consistency enhancement yields a significant performance boost, especially when the weight $\alpha$ is increased. More significantly, without the label consistency enhancement, i.e., loss $\mathcal{L}_c$, the performance decreases rapidly on SP-Motif and MNIST-75sp. Those results demonstrate the importance of label consistency enhancement in the presence of a larger distribution shift.

\subsection{Effect of Diversity Loss and Support Set Size}
In this subsection, we further investigate the effectiveness of the diversity loss $\mathcal{L}_d$ and the support set $\mathcal{S}$ in enhancing the diversity. Specifically, we alter the diversity loss weight $\alpha_d$ in Eq.~(\ref{eq:loss_d}) and the size of $\mathcal{S}$ by changing the number of graphs in each class. By default, we sample one graph from each class to ensure that there exists at least one graph in $\mathcal{S}$ sharing the same label as $G$. Denoting the average number of graphs in each class in $\mathcal{S}$ as $K$, we modify the value of $K$ to showcase the effect of $\mathcal{S}$. Note that we still ensure at least one graph in $\mathcal{S}$ sharing the same label as $G$. 

Table~\ref{tab:sens_loss} reports the performance of DLG under different values of $\alpha_d$ and $K$. From the results, we can observe that increasing the $\alpha_d$ and $K$ can generally promote the performance. The reason is that DLG  benefits from such a large diversity and thus enhances the generalizability. Nevertheless, when further increasing the value of $\alpha_d$, the performance greatly drops. This is primarily due to that the framework focuses excessively on enhancing diversity, which in contrast, diminishes the classification capability. In addition, further increasing the value of $K$ only yields marginal performance improvements. This is because these graphs are sampled from existing data, which limits additional enhancement in diversity.

%

\section{Conclusion}
In this work, we have investigated the problem of graph out-of-distribution (OOD) generalization on both graph classification and node classification. We demonstrate that the two traditional steps: augmentations and invariant subgraph extraction, may fail to achieve distribution consistency and label consistency, respectively.
To deal with this, we propose a novel framework DLG to enhance these two types of consistency. Based on learning edge masks with our designed modifiers, we are able to preserve the consistency in augmented graphs and invariant graphs. We conduct extensive experiments on a variety of both graph-level and node-level OOD generalization datasets. The results demonstrate that our framework DLG consistently outperforms other state-of-the-art graph OOD generalization methods.

\section*{Acknowledgements}
This work is supported in part by the National Science Foundation under grants (IIS-2006844, IIS-2144209, IIS-2223769, CNS-2154962, BCS-2228534, and CMMI-2411248), the Commonwealth Cyber Initiative Awards under grants (VV-1Q24-011, VV-1Q25-004), and the research gift funding from Netflix and Snap.

\bibliographystyle{IEEEtran}
\bibliography{ref}

\end{document}